\documentclass{article}

\usepackage[verbose=true,letterpaper]{geometry}
  \newgeometry{
    textheight=9in,
    textwidth=5.5in,
    top=1in,
    headheight=12pt,
    headsep=25pt,
    footskip=30pt
  }
  
\usepackage{natbib}

\usepackage{parskip}

\usepackage[utf8]{inputenc} % allow utf-8 input
\usepackage[T1]{fontenc}    % use 8-bit T1 fonts
\usepackage{hyperref}       % hyperlinks
\usepackage{url}            % simple URL typesetting
\usepackage{booktabs}       % professional-quality tables
\usepackage{amsfonts}       % blackboard math symbols
\usepackage{nicefrac}       % compact symbols for 1/2, etc.
\usepackage{microtype}      % microtypography
\usepackage{algorithmic}
\usepackage{algorithm}
\usepackage{natbib}
\usepackage{tikz}

\usepackage{xcolor}
\definecolor{dark-red}{rgb}{0.4,0.15,0.15}
\definecolor{dark-blue}{rgb}{0.15,0.15,0.4}
\definecolor{medium-blue}{rgb}{0,0,0.5}
\hypersetup{
   colorlinks, linkcolor={dark-blue},
   citecolor={dark-blue}, urlcolor={medium-blue}
}

\usepackage{amsfonts}
\usepackage{color}

\usepackage{url}
\usepackage{amsmath}
\usepackage{verbatim}

\usepackage{subcaption}

\newcommand{\pii}{$\Pi$ }
\renewcommand{\pi}{\pii }

\usepackage[flushleft]{threeparttable}
\usepackage{makecell}

\newlength{\mzerolen}

\usepackage{soul}

\usepackage{xspace}

\usepackage{upgreek}
\usepackage[export]{adjustbox}

\makeatletter
\renewenvironment{abstract}{%
    \if@twocolumn
      \section*{\abstractname}%
    \else %%
      \begin{center}%
        {\bfseries \large\abstractname\vspace{\z@}}%  %% 
      \end{center}%
      \quotation
    \fi}
    {\if@twocolumn\else\endquotation\fi}
\makeatother

\date{}

\title{The Case for Bayesian Deep Learning}

\author{
Andrew Gordon Wilson \\
\texttt{andrewgw@cims.nyu.edu} \\
Courant Institute of Mathematical Sciences \\
Center for Data Science \\
New York University \\ \\ 
\vspace{10mm}
December 30, 2019
}

\begin{document}

\maketitle

\begin{abstract}

\normalsize

\noindent The key distinguishing property of a Bayesian approach is marginalization instead of optimization, not the prior, or Bayes rule.
Bayesian inference is especially compelling for deep neural networks. (1) Neural networks are typically underspecified 
by the data, and can represent many different but high performing models corresponding to different settings of parameters, 
which is exactly when marginalization will make the biggest difference for both calibration and accuracy. (2) Deep ensembles have been 
mistaken as competing approaches to Bayesian methods, but can be seen as approximate Bayesian marginalization. 
(3) The structure of neural networks gives rise to a structured prior in function space, which reflects the inductive biases of neural networks that help them 
generalize. (4) The observed correlation between parameters in flat regions of the loss and a diversity of solutions that provide good generalization is further conducive to Bayesian marginalization,
as flat regions occupy a large volume in a high dimensional space, and each different solution will make a good contribution to a Bayesian model average. (5) Recent practical advances for Bayesian deep learning provide
improvements in accuracy and calibration compared to standard training, while retaining scalability.

\end{abstract}
\vspace{10mm}

In many situations, the predictive distribution we want to compute is given by 
\begin{align}
p(y|x, \mathcal{D}) = \int p(y|x, w) p(w|\mathcal{D}) dw \,.
\label{eqn: bma}
\end{align}
The outputs are $y$ (e.g., class labels, regression values, \dots), indexed by inputs $x$ (e.g. images, spatial locations, \dots), the weights (or parameters) of the model $f(x;w)$ are $w$, and $\mathcal{D}$ are the data. Eq.~\eqref{eqn: bma} 
represents a \emph{Bayesian model average} (BMA). Rather than bet everything on one hypothesis --- with a single setting of parameters $w$ --- we want to use every possible setting of parameters, 
weighted by their posterior probabilities. This process is called \emph{marginalization} of the parameters $w$, since the predictive distribution of interest
no longer conditions on $w$. This is not a controversial equation, but a direct expression of the sum and product rules of probability. The BMA represents 
\emph{epistemic uncertainty} --- that is, uncertainty over which setting of weights (hypothesis) is correct, given limited data. Epistemic uncertainty is sometimes 
referred to as \emph{model uncertainty}, in contrast to \emph{aleatoric} uncertainty coming from noise in the measurement process. One can naturally visualize epistemic 
uncertainty in regression, by looking at the spread of the predictive distribution as we move in $x$ space. As we move away from the data, there are many more 
functions that are consistent with our observations, and so our epistemic uncertainty should grow.

In classical training, one typically finds the \emph{regularized maximum likelihood} solution
\begin{align}
\hat{w} = \text{argmax}_w \log p(w|\mathcal{D}) = \text{argmax}_w (\log p(\mathcal{D}|w) + \log p(w) + \text{constant}).
\label{eqn: map}
\end{align}
This procedure is sometimes called \emph{maximum a-posteriori (MAP) optimization}, as it involves maximizing a posterior. 
$\log p(\mathcal{D}|w)$ is the log likelihood, formed by relating the function we want to learn $f(x;w)$ to our observations. If we are performing classification with a softmax link function, $-\log p(\mathcal{D}|w)$ corresponds to the 
cross entropy loss. If we are are performing regression with Gaussian noise, such that 
$p(\mathcal{D}|w) = \prod_{j=1}^{n} p(y_j | w, x_j) = \prod_{j=1}^{n} \mathcal{N}(y_j; f(x_i;w),\sigma^2)$, then $-\log p(\mathcal{D}|w)$ is a scaled MSE loss. In this context,
the prior $p(w)$ acts as a \emph{regularizer}. If we choose a flat prior, which has no preference for any setting of the parameters $w$ (it does not assign any feasible 
setting any more prior density than any other), then it will have no effect on the optimization solution. On the other hand, a flat prior may have a major effect on \emph{marginalization}. Indeed, even though
MAP involves a posterior and a prior, and an instantiation of Bayes rule, it is not at all Bayesian, since it is performing optimization to bet everything on a single hypothesis 
$f(x;\hat{w})$.

We can view classical training as performing approximate Bayesian inference, using the approximate posterior 
$p(w | \mathcal{D}) \approx \delta(w=\hat{w})$, where $\delta$ is a Dirac delta function that is zero everywhere 
except at $\hat{w}$.  In this case, we recover the standard predictive distribution $p(y|x,\hat{w})$. From this perspective,
many alternatives, albeit imperfect, will be preferable --- including impoverished Gaussian posterior approximations for $p(w|\mathcal{D})$, 
even if the posterior or likelihood are actually highly non-Gaussian and multimodal. 

The difference between a classical and Bayesian approach will depend on how sharply peaked the posterior $p(w|\mathcal{D})$
becomes. If the posterior is sharply peaked, there may be almost no difference, since a point mass may then be a reasonable approximation
of the posterior.
However, deep neural networks are typically very underspecified by the available data, and will thus have diffuse likelihoods $p(\mathcal{D}|w)$. Not only are
the likelihoods diffuse, but different settings of the parameters correspond to a diverse variety of compelling explanations for the data. 
Indeed, \citet{garipov2018} shows that there are large valleys in the loss landscape of neural networks, over which parameters incur 
very little loss, but give rise to high performing functions which make meaningfully different predictions on test data. \citet{zolna2019classifier}
and \citet{izmailov2019subspace}
also demonstrate the variety of good solutions that can be expressed by a neural network posterior.
This is exactly the setting when we \emph{most} want to perform a Bayesian model average, which will lead to an ensemble containing
many different but high performing models, for better accuracy and better calibration than classical training. 

The recent success of \emph{deep ensembles} \citep{lakshminarayanan2017simple} is not discouraging, but indeed strong motivation for 
following a Bayesian approach. 
Deep ensembles involves MAP training of \emph{the same architecture} many times starting from different random initializations, to find
different local optima.
Thus \emph{using these models in an ensemble is an approximate Bayesian model average}, with weights that correspond to 
models with high likelihood and diverse predictions. 
Instead of using a single point mass to approximate our posterior, as with classical training, we are now using multiple 
point masses in good locations, enabling a better approximation to the integral in Eq.~\eqref{eqn: bma} that we are trying to solve. The functional
diversity is important for a good approximation to the BMA integral, because we are summing together terms of the form $p(y|x,w)$; if two settings of 
the weights $w_i$ and $w_j$ each provide high likelihood (and consequently high posterior density), but give rise to similar models, then they will be 
largely redundant in the model average, and the second setting of parameters will not contribute much to the integral estimate. The deep ensemble 
weights can be viewed as samples from an approximate posterior \citep{gustafsson2019evaluating}. But in the context of deep ensembles, it 
is best to think of the BMA integral separately from the simple Monte Carlo approximation that is often used to approximate this integral; 
to compute an accurate predictive distribution, we do not need samples from a posterior, or even a faithful approximation to the posterior. We need
to evaluate the posterior in places that will make the greatest contributions to the integral of Eq.~\eqref{eqn: bma}.

While a recent 
report \citep{ovadia2019can} shows that deep ensembles appear to outperform some particular approaches to Bayesian neural networks, there are 
two key reasons behind these results that are actually optimistic for Bayesian approaches. First, the deep ensembles being used 
are finding many different basins of attraction, corresponding to diverse solutions, which enables \emph{a better approximation to a Bayesian model 
average} than the specific Bayesian methods considered in \citet{ovadia2019can}, which focus their modelling effort on a \emph{single}
basin of attraction. The second is that the deep ensembles require retraining a network from scratch many times, which incurs a great computational
expense. If one were to control for computation, the approaches which focus on a single basin may be preferred.

There is an important distinction between a Bayesian model average and some approaches to ensembling. The Bayesian model average
assumes that \emph{one} hypothesis (one setting of the weights) is correct, and averages over models due to an inability to distinguish 
between hypotheses given limited data \citep{minka2000bayesian}. As we observe more data, the posterior collapses, and the Bayesian model average converges
to the maximum likelihood solution. If the true explanation for the data is actually a \emph{combination} of hypotheses, the Bayesian model
average may then perform worse as we observe more data. Some ensembling methods instead work by enriching the hypothesis space, and thus 
do not collapse in this way. Deep ensembles, however, are finding different MAP or maximum likelihood solutions, corresponding to different
basins of attraction, starting from different random initializations. Therefore the deep ensemble will collapse when the posterior concentrates,
as with a Bayesian model average. Since the hypothesis space is highly expressive for a modern neural network, posterior collapse in many
cases is desirable.

Regarding priors, the prior that matters is the prior in \emph{function space}, not parameter space. In the case of a Gaussian process
\citep[e.g.][]{rasmussen2006},
a vague prior would be disastrous, as it is a prior directly in function space and would correspond to white noise. However, when we 
combine a vague prior over parameters $p(w)$ with a structured function form $f(x;w)$ such as a convolutional neural network (CNN), 
we induce a structured prior distribution over functions $p(f(x;w))$. Indeed, the inductive biases and equivariance 
constraints in such models is why they work well in classical settings. We can sample from this induced prior over functions by first sampling parameters 
from $p(w)$ and then conditioning on these parameters in $f(x;w)$ to form a sample from $p(f(x;w))$ \citep[e.g.,][Ch 2]{wilson2014thesis}.
Alternatively, we can use a neural network kernel with a Gaussian process, to induce a structured distribution over functions \citep{wilson2016deep}.

Bayesian or not, the prior, just like the functional form of a model, or the likelihood, will certainly be imperfect, 
and making unassailable assumptions will be impossible. Attempting to avoid an important part of the modelling process
because one has to make assumptions, however, will often be a worse alternative than an imperfect assumption. 
There are many considerations one might have in 
selecting a prior. Sometimes a consideration is invariance under reparametrization.
Parametrization invariance is also a question in considering
regularizers, optimization procedures, model specification, etc., and is not specific to 
whether or not one should follow a Bayesian approach. Nonetheless, we will make some
brief additional remarks on these questions.

If we truly have a vague prior over parameters, perhaps subject to some constraint for normalization, then our posterior reflects essentially
the same relative preferences between models as our likelihood, for it is a likelihood scaled by a factor that does not depend on $w$ outside some 
basic constraints.
In computing the integral for a Bayesian model average, each setting of parameters is weighted by the quality of the associated function, as measured
by the likelihood. Thus the model average is happening in function space, and is invariant to reparametrization. In the context of many standard 
architectural specifications, 
there are also some additional 
benefits to using relatively broad zero-mean centred Gaussian priors, which help provide smoothness in function space by bounding the norm of the weights. But this smoothness is not 
a central reason to follow a Bayesian approach, as one could realize similar advantages in performing MAP 
optimization. Bayesian methods are fundamentally about marginalization as an alternative to optimization. 

Moreover, vague priors over parameters are also often a reasonable description of our a priori subjective beliefs. We want to use a given functional form, which is 
by no means vague, but we often do not
have any strong a priori preference for a setting of the parameters. 
It is worth 
reiterating that a vague prior in parameter space combined with a highly structured model such as a convolutional neural network does \emph{not} 
imply a vague prior in function space, which is also why classical training of neural networks provides good results.
Indeed, vague parameter priors are often preferable to entirely ignoring 
epistemic uncertainty, which would be the standard alternative.
In fact, ignoring epistemic uncertainty is a key reason that standard neural network training is \emph{miscalibrated}. By erroneously
assuming that the model (parameter setting we want to use) is completely determined by a finite dataset, the predictive distribution becomes \emph{overconfident}: for example, the highest softmax
output of a CNN that has undergone standard training (e.g. MAP optimization) will typically be much higher than the probability of the corresponding class label 
\citep{guo2017calibration}.
Importantly, ignoring epistemic uncertainty also leads to worse accuracy in point predictions, because we are now ignoring all the other compelling explanations
for the data. While improvements in calibration are an empirically recognized benefit of a Bayesian approach, the enormous potential for 
gains in \emph{accuracy} through Bayesian marginalization with neural networks is a largely overlooked advantage. 

There are also many examples where flat priors over parameters 
combined with \emph{marginalization} sidestep pathologies of maximum likelihood training. 
Priors without marginalization are 
simply regularization, but Bayesian methods are not about regularization \citep[][Ch 28]{mackay2003information}. 
And there is a large body of work considering approximate Bayesian 
methods with uninformative priors over parameters (\emph{but not functions}) 
\citep[e.g.,][]{clyde2004model, berger1996intrinsic, ohagan1995fractional, berger2006case, gelman2013bayesian, mackay2003information, mackay1992, neal1996}. 
This approach is well-motivated, marginalization is still compelling, and the results are often better than regularized optimization.

The ability for neural networks to fit many datasets, including images with random labels \citep{zhang2018understanding}, is indicative of their support, rather 
than their inductive biases. We want to have a large support, because we believe a priori that there are many possible explanations for a given problem,
even if a large fraction of these explanations are a priori improbable. The distribution of this support --- which solutions are a priori probable --- is determined
by the inductive biases of the model \citep[e.g.,][]{wilson2014thesis}. Gaussian processes with popular kernels are capable of expressing large support, with near universal approximation properties, but at the same time very simple inductive biases \citep{rasmussen2006}. A Gaussian process can also perfectly fit noise, but favours more structured solutions. 
Similarly, a vague distribution over parameters in a neural network induces a distribution over functions that has support for many solutions, as we would hope.
But neural networks have been constructed not just for flexibility, but to have inductive biases such that the mass of good solutions would reasonably exceed the mass of bad solutions for problems that neural networks are typically applied to. Indeed, flexibility (aka support) can be achieved purely through \emph{width}, but \emph{depth} is what has led to great improvements
in generalization. 

Moreover, it has been observed that solutions in flat regions of the loss surface correspond to better generalization \citep{keskar2016large, hochreiter1997flat, izmailov2018}. Although flatness can be achieved through reparametrization \citep{dinh2017sharp}, the flatness that is associated with good generalization is when parameters in the flat region correspond to different high performing models \citep{garipov2018, izmailov2018}. In a high 
dimensional space, these solutions in flat regions will take up much more volume than bad solutions \citep{huang2019understanding}, even if they have the same values of the loss, which is partly why they are easily discoverable by optimization procedures. This observation further motivates the Bayesian approach to deep learning: with a focus on \emph{integration} rather than \emph{optimization}, the high volume solutions that provide good generalization and are associated with functional diversity (avoiding redundancy in the model average) will have a much greater effect on the predictive distribution than the sharper solutions.

Indeed, by accounting for epistemic uncertainty through uninformative \emph{parameter} (but not function) priors, we, as a community, have developed 
Bayesian deep learning methods with improved calibration, reliable predictive distributions, and improved accuracy 
\citep[e.g.,][]{mackay1992bayesian, neal1996, gal2016dropout, saatci2017bayesian, kendall2017uncertainties, ritter2018scalable, khan2018fast, maddoxfast2019, sun2019functional, izmailov2019subspace, zhang2019cyclical}. \citet{mackay1995probable} and \citet{neal1996} are particularly notable early works considering
Bayesian inference for neural networks. \citet{seeger2006bayesian} also provides a clear tutorial on Bayesian methods in machine learning.
Of course, we can always make better assumptions --- Bayesian or not. We should strive to build more interpretable parameter priors. There 
are works that consider building more informative parameter priors for neural networks by reasoning in function space \citep[e.g.,][]{sun2019functional, yang2019output, louizos2019functional, hafner2018reliable}.
And we should also build better posterior approximations. Deep ensembles are a promising step in this direction. 

But we should not undermine the progress we are making so far. Bayesian inference is especially compelling for deep neural networks. Bayesian deep learning 
is gaining visibility because we are making progress, with good and increasingly scalable practical results. We should not discourage these efforts.
If we are shying away from an approximate Bayesian approach because of some challenge or imperfection, we should always ask, \emph{``what is the alternative''?} The alternative may 
indeed be a more impoverished representation of the predictive distribution we want to compute.

There are certainly many challenges to computing the integral of Eq.~\eqref{eqn: bma} for modern neural networks, including a posterior landscape which is 
difficult to navigate, and an enormously high (e.g., 30 million) dimensional parameter space. Many of the above papers are working towards addressing such 
challenges.
 We have been particularly working on 
recycling geometric information in the SGD trajectory for scalable approximate Bayesian inference \citep{izmailov2019subspace, maddoxfast2019}, exploiting large loss valleys \citep{garipov2018}, and creating subspaces of low
dimensionality that capture much of the variability of the network \citep{izmailov2019subspace}. \citet{pradier2018latent} also considers different approaches
to dimensionality reduction, based on non-linear transformations. For exploring multiple distinct basins of attraction, we have been developing cyclical stochastic MCMC approaches \citep{zhang2019cyclical}, which could be seen as sharing many of the advantages of deep ensembles, but with an added attempt to also marginalize 
within basins of attraction.

\normalsize{
\normalsize{
\bibliographystyle{apalike}
\bibliography{mbibnew}
}
}

\end{document}